\documentclass{article}
\usepackage{spconf,amsmath,graphicx,subfigure}

\title{CARDIAC MOTION ANALYSIS BY TEMPORAL FLOW GRAPHS}

\name{V S R Veeravasarapu$^*$\thanks{*The work was done when this author is working at CVIT, IIIT-Hyderabad.} \qquad Jayanthi Sivaswamy$^{1}$ \qquad Vishanji Karani$^{2}$}
 \address{$^{1}$ Center for Visual Information Technology, IIIT-Hyderabad, INDIA \\
     $^{2}$CARE Hospitals,	Hyderabad, INDIA\\
     }
\begin{document}
\ninept
\maketitle
\begin{abstract}
Cardiac motion analysis from B-mode ultrasound sequence is a key task in assessing the health of the heart. The paper proposes a new methodology for cardiac motion analysis based on the temporal behaviour of points of interest on the myocardium. We define a new signal called the \emph{Temporal Flow Graph} (TFG) which depicts the movement of a point of interest over time. It is a graphical representation derived from a flow field and describes the temporal evolution of a point. We prove that TFG for an object undergoing periodic motion is also periodic. This principle can be utilized to derive both global and local information from a given sequence. We demonstrate this for detecting motion irregularities at the sequence, as well as regional levels on real and synthetic data. A coarse localisation of anatomical landmarks such as centres of left/right cavities and valve points is also demonstrated using TFGs.
\end{abstract}
\section{Introduction} \label{sec:intro}


Myocardial motion analysis from a series of B-mode ultrasound (US) images has received attention in the past decade and still remains an open and challenging problem. It broadly aims at obtaining information about the location and extent of any damage to heart tissue. Some of the specific information of interest are beat intervals and locations (heart rate variability); missing of beats; behaviour of anatomical landmarks (the apex, the mitral annulus and center of the left and right ventricle cavities). Generally, cardiologists get these information by manual inspection though, attempts have been made to develop tools which assist via automatic segmentation of major anatomical structures like left ventricle (LV) and analysing the changes in their shape or volume \cite{qazi2007automated} etc. 

Tools for a global assessment of the cardiac motion are of also of interest since they facilitate detecting the overall health of the heart. Automatic 3D cyclic motion analysis has been addressed for general data in \cite{seitz1997cyclic}. A function based on the temporal correlation of frames (called``Period Trace") was defined to help profile the motion in a given sequence in terms of changes in the period. This function is very sensitive to noise and hence, is not suitable to analyse motion trends and irregularities in US sequences. Learning-based approach is proposed for discriminating between normal and abnormal motion in \cite{syeda2006characterizing}. Average velocity curves for regions of interest are derived by analysing regional motion trends between image sequences. Such regions of interest are first segmented from the image a priori. These curves are learned from a set of image sequences and used for discrimination between normal and abnormal motion. This method was further improved by incorporating demon's algorithm to derive robust motion fields and using a support vector machine classifier to achieve higher classification accuracy \cite{syeda2007characterizing} and \cite{wang2008spatio}. However, the accuracy of the methods heavily depends on the segmentation which is a challenging task in US images. Another method for detecting abnormalities is based on estimating the motion field using a moving spatiotemporal B-spline window \cite{suhling2005myocardial}. The method poses the motion parameter estimation as a least squares estimation problem and hence is computationally very intensive.

We proposed a concept we call temporal flow graph (TFG) for cardiac cyclic motion analysis of a myocardial point from B mode US data. TFG is a graphical representation of the evolution of an interest point throughout the course of a motion cycle. In this paper, we present this concept and its scope of applicability for a range of applications. We demonstrate its ability to support both a global and local assessment of cardiac motion.  


The remainder of the paper is organized as follows. In section \ref{sec:TFG}, we introduce the concept of TFG. In section \ref{scopeTFG}, we discuss about two applications of TFG which are important tasks in cardiac asessment. We conclude with some future directions of our work in the section\ref{sec:conc}.

\section{\uppercase{Temporal Flow Graphs}}   \label{sec:TFG}
\begin{figure}
\centering
\subfigure[]{\includegraphics[width=3.7cm]{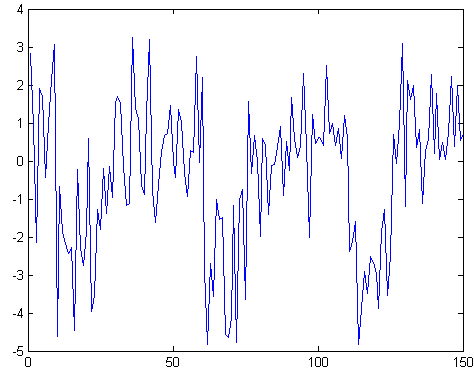}} 
\subfigure[]{\includegraphics[width=3.8cm]{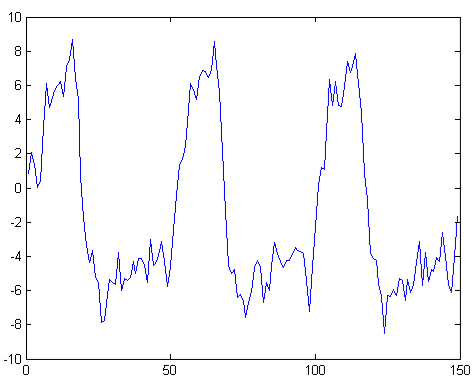}}
\caption{(a) IDG and (b) TFG graphs for a landmark}
\label{tfg}
\end{figure}

Consider a point of interest $p$ undergoing motion captured in an video sequence. The displacement/motion of this point over two frames is represented by a vector $v$. Estimation of such a vector at every point over the entire sequence yields a dense motion field commonly referred to as optical flow. Let us consider quantising the direction (or angle, $\theta$) of the motion vector to be +1 when $\theta < 180^0$ and -1 when $\theta > 180^0$. This helps represent the displacement of $p$ over time as a 1-D function of time where the value of the function is determined by the displacement magnitude and the sign is determined by the displacement direction. We call such a function as a Instantaneous Displacement Graph (IDG).  
Let $I$ be a given image sequence and let the displacement at a point of interest $p$ between the $n^{th}$ and $(n+1)^{th}$ frame be of magnitude $v_n$ in the direction $\theta_n$. We first quantise the direction as follows

\begin{equation*}
dir(\theta_n) =
\begin{cases}
 -1, &  -180^0\le \theta_n < 0^0 \\
 +1, &    0^0\le \theta_n < 180^0
\end{cases}
\end{equation*}
Then we define the IDG for the point $p$ as follows
\begin{equation}
\label{idg}
IDG_p(n) =  v_n.dir(\theta_n)
\end{equation}
The cumulative sum of IDG helps track the motion flow of a point over time. We refer to such a sum as a Temporal Flow Graph (TFG). Such graphs helps represent the cyclic motion history of a point in a compact manner. Mathematically, the TFG is defined as
 
\begin{equation}
TFG_p(k) = \sum_{n=1}^k IDG_p(n)                
\end{equation}
where $k$ is the frame index. Fig.\ref{tfg} shows a sample of IDG and TFG of a point on myocardial boundary on cardiac sequence. When the point $p$ undergoes cyclic motion, the periodicity is readily captured by the TFG rather than the IDG. This is stated as a Lemma below:

\noindent\textbf{Lemma:} \emph{The running sum of the instantaneous displacement of a point undergoing cyclic motion is a periodic function. } \\
\textbf{Proof :} \\
\indent Without loss of generality, consider a sequence in which a simple pendulum oscillates about a equilibrium point ($O$). The point of interest is the centre of mass ($P$). Let $x_n$ and $\theta_n$ be the instantaneous distance and angular displacements that the mass undergoes between $n^{th}$ and $(n+1)^{th}$ frames. Let $X$ and $\theta$ be the distance and angular displacement of $P$ from $O$ respectively.  
The equation of motion for the simple pendulum for sufficiently small amplitude has the form as shown below.
\begin{equation*}
\frac{d^2X}{dt^2}+\frac{g}{L}X=0
\end{equation*}
where $g$ is the gravitational constant, $L$ is the length of pendulum and $t$ refers to frame number or time-line. The assumption here is that the frame rate is sufficiently large for capturing the cyclic motion (at least two times the highest frequency in the cyclic motion). The solution for the above differential equation is of the form
\begin{equation*}
X(t)=A cos(wt+\phi)
\end{equation*}
\begin{equation} \label{property}
But, \ \ \ \ 
X(t)= \sum_{n=1}^t x_n = \sum_{n=1}^t L.\theta_n = A cos(wt+\phi)
\end{equation}
In Eq.\ref{property}, $X(t)$ is nothing but $TFG$ and $x_n$ is nothing but $IDG$. 
\begin{equation}
TFG_P(t)= \sum_{n=1}^t IDG_P(n) = A cos(wt+\phi)
\end{equation}
Hence, TFG is periodic. Its period is $\frac{2\pi}{w}$ which is identical to the period of cyclic motion of pendulum.

\subsection{Optical Flow Estimation} \label{opticalflow}
The parameters $(v_n, \theta_n)$ in Eq.\ref{idg} can be derived from the optical flow or the motion field of estimated from two consecutive frames. Optical flow field is computed under the brightness constancy assumption. This leads to the well known optical flow equation:




 \begin{equation}
 I_x v_x + I_y v_y + I_t = 0
 \end{equation}

 where $v_x$ and $v_y$ are the projections of the velocity vector along the $x$ and $y$ axes. The brightness constancy assumption is very sensitive to brightness changes. Therefore, it is relaxed and a gradient constancy assumption has been proposed \cite{brox2004high} which is expressed as:

 \begin{equation}
 \nabla I (x, y, t) = \nabla I (x + v_x, y + v_y,t +1)
 \end{equation}

 Velocity vectors $v_x$ and $v_y$ are determined by minimizing the total of energy $E(v_x,v_y)$. 




 Additionally, a smoothness assumption is also imposed on the flow field. Further details on this can be found in \cite{mailloux1989restoration}. We follow this method and compute the optical flow.


 The desired flow parameters between $n^{th}$ and $(n+1)^{th}$ frames are found from the flow field as 
 \begin{equation}
 v_n = \sqrt{v_x^2 + v_y^2}
 \end{equation}

 \begin{equation}
 \theta _n = \tan ^{-1} (\frac{v_y}{v_x})
 \end{equation}

Next, we showcase the utility of the TFG with an application which is of key interest in medical imaging. 

\section{\uppercase{Scope of Temporal Flow Graphs}} \label{scopeTFG}
We consider three categories of problems: i) abnormal motion detection such as missing beats and beat-pause, ii) abnormal region detection and iii) anatomical landmark detection. We use both simulated and real ultrasound data for our experimentation. Simulated data was generated by using the Ultrasound simulation package in \cite{yu2002speckle}.

\subsection{Abnormality detection}
A major task in cardiac motion analysis is abnormality detection. Abnormalities can be either in motion or in a region. Examples of the former are Tachycardia, missing beat, and beat pause etc. Classification of cardiac motion as normal/abnormal has been attempted by learning on Bayesian networks of LV volume changes \cite{qazi2007automated}. Examples of abnormality at the regional level are coronary heart disease, valvular heart disease, etc. 
Disease type and its extent is generally decided based on the location of abnormal region. We show that both types of abnormalities can be detected using the TFG.

\subsubsection{Abnormal motion detection} \label{ad}
Missing beats often occurs in functional cardiovascular disease, Hyperthyroidism and myocardial disease. A "beat-pause" for morethan 4  seconds is considered as ventricular arrest. It may occur in patients who suffer from functional cardiovascular disease, such as myocarditis, myocardial infarctions, cardiac pathological changes, etc. 

\begin{figure}[h!]
\centering
\subfigure[TFG]{\includegraphics[height=2.5cm]{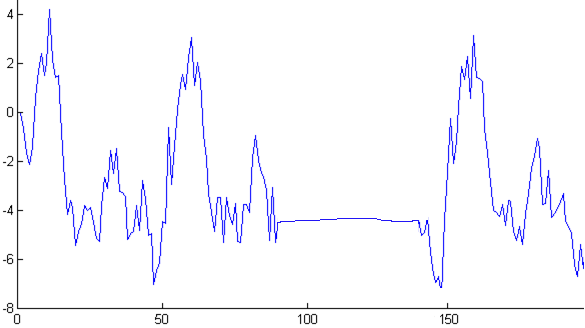}} \label{am_a}
\subfigure[Short Time Variance plot]{\includegraphics[height=2.5cm]{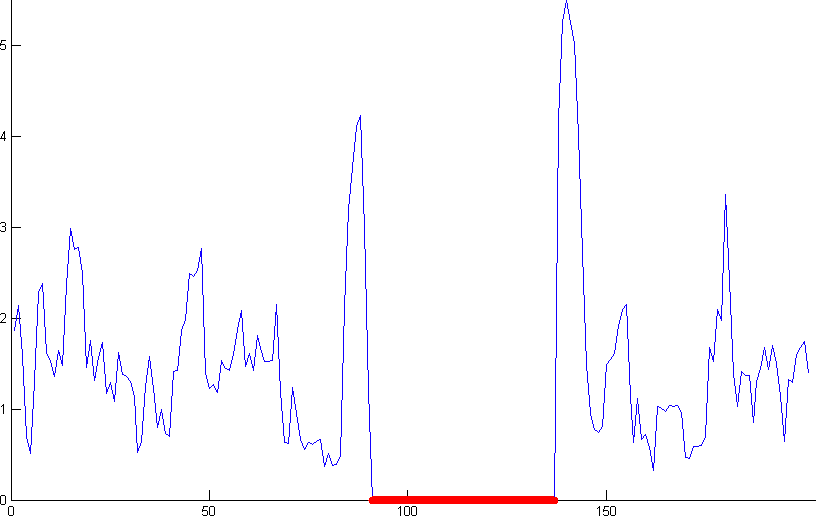}} \label{am_b}
\caption{Abnormal motion detection}
\label{abnormal_motion}
\end{figure}

Abnormal motions can be detected by a local analysis on TFGs, such as energy content in the signal in short time intervals. For this experiment, a synthetic US image sequence was generated in which a beat-pause was introduced in 58 frames ($\sim$ 2 seconds). The myocardial point on this sequence was automatically detected first using a series of morphological operations such as smoothing, binarization, closing and fill holes. Next the TFG was computed for this myocardial point and is as shown in Fig.\ref{abnormal_motion}(a). The length of the beat-pause is detected by examining the signal variance associated with beat-pulse region. This is larger in normal regions compared to the beat-pause region as it does not have any energy. Hence, a short-time-variance was used for this purpose. Overlapping windows of size 8 samples and shift by one sample were used for short-time-variance computation. The short-time-variance plot for the TFG shown in Fig.\ref{abnormal_motion}(a) is as shown in Fig.\ref{abnormal_motion}(b). Next, beat-pause region is detected (annotated with red color in Fig.\ref{abnormal_motion}(b)) by a threshold T(=0.2 in our case). The detected beat-pause length is 54 samples (ground truth is 58) i.e. 1.86 seconds. Hence, this is classified as \textit{missing beats}. If the duration of the beat pause in the TFG graph is more than 4 seconds, then this is deemed to be due to ventricular arrest. By using a similar approach, heart rate and its variations can also be analysed. 

\subsubsection{Abnormal region detection} \label{ard}
\begin{figure}[h!]
\centering
\subfigure[]{\includegraphics[height=1.4cm]{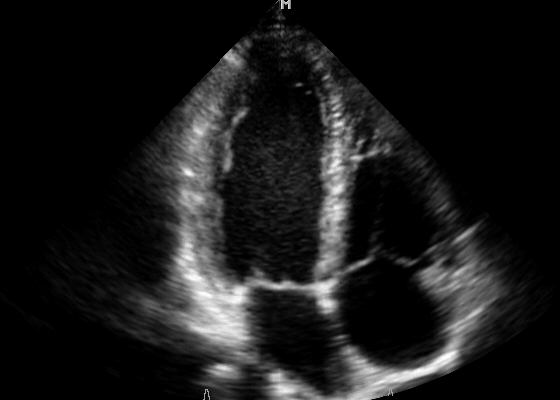}} 
\subfigure[]{\includegraphics[height=1.4cm]{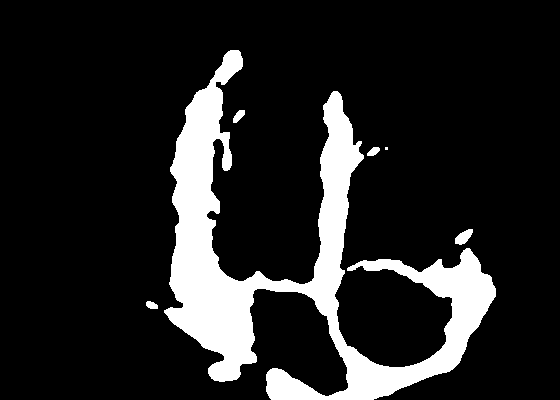}}
\subfigure[]{\includegraphics[height=1.4cm]{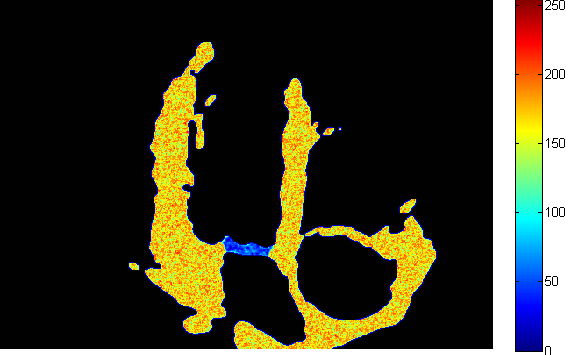}}
\subfigure[]{\includegraphics[height=1.4cm]{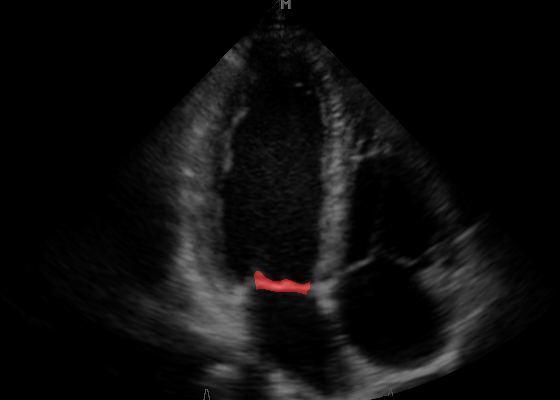}}
\caption{Abnormal heart muscle detection results: (a)First frame; (b)detected myocardium; (c)pseudo-coloured  variance map; (d) detected abnormal region overlaid on the first frame.}
\label{dead}
\end{figure}

Detection of dead/dying heart muscle is a key and preliminary task in diagnosis of many heart diseases such coronary artery disease, rheumatic heart disease, valvular heart disease etc. Cardiologists visually inspect a US sequence to observe the motion of muscle, for diagnosis. Here, we propose a way to automate the process of locating the dead muscle based on the fact that dead/dying heart muscle does not move \textit{coherently} with myocardium. For an healthy heart, all points on the myocardium move coherently (same cyclic behaviour) even-though their instantaneous displacements are different. This fact is reflected in the variance map of the TFGs of pixels on myocardium. 
Hence, an abnormal region can be detected by simple thresholding on the variance map. A synthetic US sequence was generated using \cite{karavides2010database} in which tricupsid valve is not moving. This was done in the following steps: a) Detection of myocardium as described earlier; (b) Generation of TFGs for all myocardial points; (c)computation of the variance map; and finally (d) Thresholding of the variance map. Fig.\ref{dead} shows the first frame, detected myocardium, color coded variance map and abnormal region (marked in red color).

\subsection{Anatomical Landmarks detection} \label{ld} 

Landmarks on ultrasound (US) images are required for point-based registration applications \cite{makela2002review}, \cite{loats1993ct}. These methods often use external markers or manual detection of anatomical landmarks. But automatic detection of landmarks is of growing interest in general and for echocardiography, several methods have been proposed in past decade. Van Stralen et al.\cite{van2008time} proposed a method to detect left ventricular long axis and mitral valve location by using a circular Hough transform and dynamic programming. Classification-based approach has also been attempted for the detection. A cascade of three classifiers is trained based on boosting techniques using Haar wavelet-like features and steerable features to detect standard view planes in \cite{lu2008autompr}. Standard anatomical landmark points (apex, mitral valve points) in two-chamber (2C) and four chamber (4C) view planes have also been detected using a classification based method \cite{karavides2010database}.

\begin{figure}
\centering
{\includegraphics[height=2.2cm]{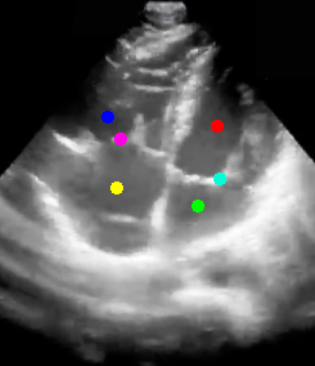}}
{\includegraphics[height=2.2cm]{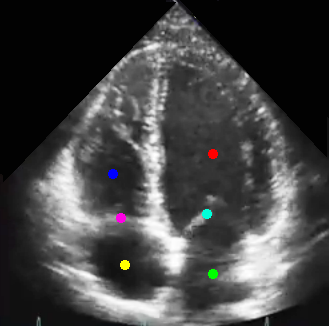}}
{\includegraphics[height=2.2cm]{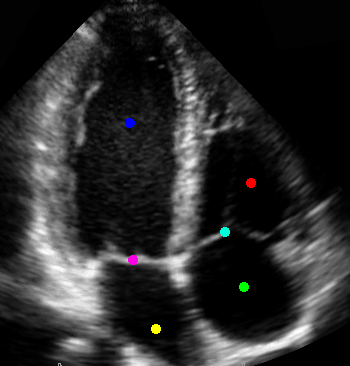}}
{\includegraphics[height=2.2cm]{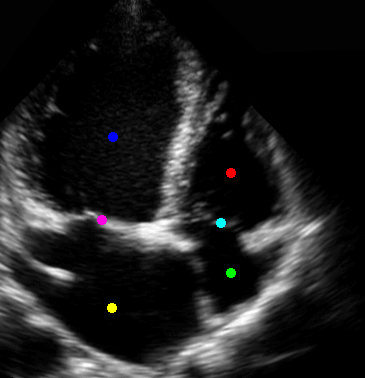}}
\caption{Anatomical Landmarks detection: Positions of landmarks on first frames of 4 US sequences. Left ventricle center (red), Right ventricle center (blue), Left atrium center (green), Right atrium center (yellow), Mitral valve point (cyan), and Tricupsid valve point (magenta).}
\label{Landmark}
\end{figure}

Here, we propose an automated process of detection of six anatomical landmarks based on their temporal flow using the TFG concept. Tricuspid and mitral valve points undergo rapid motion and hence their TFGs should have maximum variance. On the other hand, the centres of left ventricle (LV), right ventricle (RV), left atrium (LA) and right atrium (RA) cavities do not undergo significant motion and therefore the corresponding TFGs should have minimum variance. These principles and the domain knowledge about their spatial relationship in 4C view planes can be used in detecting these landmarks. We propose detecting these landmarks on ultrasound image sequences through two steps: i) TFG computation for all pixels in the first frame and ii) extrema computation on the variance of TFGs for the pixels in the right, left, top and bottom quadrants of the image. The proposed approach was evaluated on a dataset which contains 4C view-US sequences recorded on 19 patients. The positions of those landmarks on first frames of different US sequences are shown in the Fig.\ref{Landmark}. The red, blue, yellow and green colored landmarks are the points with minimum TFG variance in different quadrants. These are the desired centres of LV, RV, RA and LA cavities. The cyan and magenta coloured landmarks are points of maxima in the TFG variance. These are points on heart valves (mitral and tricuspid). The results were assessed against the ground-truth marked by an expert. Corresponding location errors are summarized in Table \ref{table_location}.

\begin{table}[h!]
\begin{center} 
\caption{Average location errors on 4C view of 19 US sequences}
  \begin{tabular}{ | c || c | c |}
  \hline
  Landmark & mean (in mm) & std (in mm) \\
   \hline \hline
  Mitral valve point & 7.2 & 3.7  \\
  \hline
  Tricupsid valve point & 5.1 & 5.9  \\ 
  \hline
  Left ventricle cavity center & 10.9 & 4.1  \\ 
  \hline 
  Right ventricle cavity center & 9.6 & 6.8 \\ 
  \hline
  Left atrium cavity center & 7.5 & 5.2 \\
  \hline
  Right atrium cavity center & 6.7 & 3.0 \\
  \hline
  \end{tabular}
  
\label{table_location}
\end{center}
\end{table}

We compare the performance of our approach to some published approaches which identify the mitral valve point. \cite{van2008time} detects the mitral valve ring and has larger errors than our method (Table 2). However, other methods \cite{orderud7265automatic}, \cite{esther2008sparse}, \cite{lu2008autompr}, and \cite{karavides2010database} outperform our method. It should be pointed out that the proposed method was aimed only at finding the approximate location of the landmark using a simple technique such as the temporal behaviour. In contrast, \cite{lu2008autompr} achieves the lowest error at a much higher computational cost after training on 244 cases. 
Another interesting departure in the approaches is that the proposed method is able to detect six landmarks simultaneously whereas most existing work use different strategies to locate different landmarks. The output of our method can be used as a seed for a stage which localises the landmarks more accurately. The fact that errors of the proposed method and that reported in \cite{van2008time} are close is encouraging. It also suggests that our method is accurate enough to replace the manual interaction. 

\begin{table}[h!]
\begin{center} 
\caption{Comparison on MV landmark detection}
  \begin{tabular}{ | c || c | c |}
  \hline
  Method & mean(in mm) & std (in mm) \\
   \hline \hline
  Orderud et al. \cite{orderud7265automatic} & 3.6 & 1.8 \\
  \hline
  Lu et al. \cite{lu2008autompr} & 3.6 & 3.1 \\ 
  \hline
  Leung et al. \cite{esther2008sparse} & 4.5 & 2.9  \\ 
  \hline 
  Karavides et al. \cite{karavides2010database}& 5.0 & 2.5 \\
  \hline
  Van Stralen et al. \cite{van2008time} & 8.4 & 5.7  \\ 
  \hline
  Proposed& 7.2 & 3.7  \\
  \hline  
  \end{tabular}
  
\label{table_compare}
\end{center}
\end{table}

\section{\uppercase{conclusions}} \label{sec:conc}
A new methodology has been presented for cardiac motion analysis based on the temporal behaviour of some points of interest on the myocardium. A new signal called the \emph{Temporal Flow Graph} which describes the temporal behaviour of the landmark has been defined. Abnormality detection and anatomical landmark detection are shown to be simplified using the TFG. The proposed method for abnormal region detection (section \ref{ard}) may not work for the case of dying heart muscle which has abnormal motion, but same periodicity. However, training based and feature classification methods can be helpful in such scenario. 

The complexity of TFG production lies in the optical flow (OF) computation stage. Thus, real-time implementation of TFG-based cardiac motion analysis is only limited by the choice of the OF algorithm. The proposed algorithms can also be extended for in-vivo analysis for animal studies and for assessing fetal cardiovascular health. 

\bibliographystyle{IEEEbib}
\bibliography{refs}

\end{document}